\documentclass{article}

\usepackage[final]{corl_2019} 

\usepackage[utf8]{inputenc}
\usepackage{amsmath}
\usepackage{amsfonts}
\usepackage{pgfplots}
\usepackage{graphicx}
\usepackage{subcaption}
\usepackage{algorithm2e}
\usepackage{array}
\usepackage{makecell}

\newcommand{\cvec}[1]{\boldsymbol{\mathrm{#1}}}
\newcommand{\cmat}[1]{\boldsymbol{\mathrm{#1}}}
\newcommand\given[1][]{\:#1\vert\:}

\title{Receding Horizon Curiosity}

\author{Matthias Schultheis, Boris Belousov, Hany Abdulsamad, Jan Peters$^\dagger$\\
	Department of Computer Science, Technische Universit\"at Darmstadt, Germany\\
	$^\dagger$Robot Learning Group, Max Planck Institute for Intelligent Systems, T\"ubingen, Germany\\
	\small\texttt{matthias.schultheis@gmail.com},
	\texttt{\{belousov, abdulsamad, peters\}@ias.tu-darmstadt.de}
}

\begin{document}
\maketitle


\begin{abstract}
    Sample-efficient exploration is crucial not only for discovering rewarding experiences
    but also for adapting to environment changes in a task-agnostic fashion.
    A principled treatment of the problem of optimal input synthesis for system identification
    is provided within the framework of sequential Bayesian experimental design.
    In this paper, we present an effective trajectory-optimization-based
    approximate solution of this otherwise intractable problem
    that models optimal exploration in an unknown Markov decision process (MDP).
    By interleaving episodic exploration with Bayesian nonlinear system identification,
    our algorithm takes advantage of the inductive bias to explore in a directed manner,
    without assuming prior knowledge of the MDP.
	Empirical evaluations indicate a clear advantage of the proposed algorithm
	in terms of the rate of convergence and the final model fidelity
    when compared to intrinsic-motivation-based algorithms
    employing exploration bonuses such as prediction error and information gain.
    Moreover, our method maintains a computational advantage
    over a recent model-based active exploration (MAX) algorithm,
	by focusing on the information gain along trajectories
    instead of seeking a global exploration policy.
    A reference implementation of our algorithm and the conducted experiments
    is publicly available\footnote{\url{https://github.com/mschulth/rhc}}.
\end{abstract}



\section{Introduction}
Learning agents are expected to constantly adapt to changing environments
without an otherwise explicit task specification.
A subclass of such changes relates to predicting the consequences of the agent's own actions,
as captured by the forward model of the environment,
which is of crucial importance for the purposes of planning and task-solving.
When the forward model parameters are unknown,
what sequence of actions should the agent take to reveal the most about the system,
i.e., what course of action facilitates the most sample-efficient exploration?
To answer this question, the information content in the observations needs to be quantified,
which becomes possible if the agent maintains a probability distribution over the model parameters.
The sequential decision making problem with such a probabilistic model whose parameters are not directly observable
is known as the partially observable Markov decision process (POMDP)~\cite{kaelbling1998planning}.

Finding an exact solution of continuous-state/action POMDPs is intractable in general~\cite{papadimitriou1987complexity}.
Therefore, various approximations are commonly employed~\cite{ghavamzadeh2015bayesian}.
In particular, one-step greedy heuristics---exploration bonuses---have gained
prominence recently due to the ease of implementation,
wide range of applicability, and good empirical performance in game domains~\cite{burda2018large}.
However, they neglect long-term effects of control actions
and therefore struggle in underactuated continuous control domains,
focusing on immediate curiosity satisfaction~\cite{shyam2018model}.
As a remedy, it was proposed that the agent should plan exploration~\cite{sun2011planning},
purposefully driving the system to the states that provide the largest information gain,
thus performing active exploration.

Many active exploration approaches have been proposed
in the past~\mbox{\cite{mehra1974optimal,gevers2006input,zarrop1979optimal,huan2016sequential,klenske2016dual,ling2016gaussian}},
differing chiefly in how the model is represented and how it is used for planning.
The most recent representative from this family of approaches
is the model-based active exploration (MAX) algorithm~\cite{shyam2018model},
that represents the uncertainty in the dynamics via an ensemble of deterministic networks,
and uses a model-free reinforcement learning (RL) algorithm
to find an exploration policy optimal with respect to the ensemble.
Such approach is computationally demanding because it solves a full RL problem in the inner loop
to only use the optimized policy for performing a few exploratory steps.

In this paper, we present a principled model-based active exploration method,
which contrasts with the recent computationally expensive RL-based approaches.
The proposed exploration algorithm represents the uncertainty in the model
by a distribution over the parameters in a shallow Bayesian network
and finds optimally explorative actions via trajectory optimization based on the learned model.
Since rigid-body dynamics can be written as a dot product between
a vector of state-action features and a vector of physics parameters,
we adopt a similar model structure with generic feature functions and a Gaussian vector of unknown parameters.
Due to the linear-Gaussian structure, belief space dynamics can then be obtained in closed-form
and incorporated into the trajectory optimization formulation,
capturing the effect of current actions on future information gain.

The proposed method, termed receding horizon curiosity (RHC),
addresses several challenges involved in the design and implementation of actively exploring agents.
First, the agent's beliefs must be represented and propagated in time to estimate the information gain.
A combination of approximations is needed and it is not clear a priori if the resulting algorithm will work.
RHC employs Gaussian beliefs and a maximum likelihood observation assumption to represent and propagate the beliefs.
Second, the information gain objective needs to be evaluated and optimized.
Again, a number of approximations are involved, that require empirical evaluation.
RHC exploits the Gaussian-linear model structure to evaluate the information gain
and it relies on trajectory optimization to maximize it.
Finally, a key feature of RHC is the interleaved optimal exploration and model updating,
which turns out to be sufficient for promoting efficient model-learning, as shown by the experimental evaluations.
On the whole, RHC compares favorably with state-of-the-art model-free intrinsic motivation approaches
in terms of the model error and downstream task performance in classical continuous control environments,
and compared to MAX, it is computationally far less demanding and has lower variance over runs.

\section{Foundations}
\label{sec:foundations}

In this section, the background on active learning, Bayesian linear regression,
random Fourier features, model-based reinforcement learning, and multiple shooting methods is provided.

\subsection{Active Learning}
\label{subsec:active-learning}
In supervised learning, a training set is predefined and fixed.
However, if an agent is allowed to choose the instances to train on,
learning can potentially progress faster and require fewer samples.
Active learning~\cite{settles2009active},
also known as optimal experiment design~\cite{zarrop1979optimal},
is an area of statistical learning that addresses exactly the question of
how to choose the data points for learning.
Choosing an optimal subset of points involves a combinatorial number of possibilities.
Therefore, approximations of the optimal value function,
known as \emph{acquisition functions}, are employed for query point selection.

Perhaps the most straightforward and common query framework
is~\emph{uncertainty sampling}~\cite{settles2009active}.
In this framework, an active learner queries the instance $x\in X$
for which the model output $y\in Y$ is least certain.
With the entropy of the output $\mathbb{H}(y | x)$ as the measure of uncertainty,
the following acquisition function needs to be maximized
\begin{align*}
    \alpha_\textrm{us}(x) = \mathbb{H}(y | x).
\end{align*}
Compared to other approaches, uncertainty sampling is computationally relatively light,
and if the likelihood belongs to an exponential family,
the entropy can even be obtained in closed-form.
A drawback of uncertainty sampling is that it fails if the underlying system is stochastic,
as it cannot distinguish between aleatoric and epistemic uncertainty.

To fix the shortcomings of uncertainty sampling,
\emph{expected variance reduction}~\cite{settles2009active} was proposed,
that explicitly takes into account the model variance, by minimizing the acquisition function
\begin{align*}
	\alpha_\textrm{evr}(x) =
    \textrm{var} \left( \mathcal{M} \given \mathcal{D} \cup (x, y) \right),
\end{align*}
where $\textrm{var} \left( \mathcal{M} \given \mathcal{D} \cup  (x, y) \right)$
denotes a measure of variance of the model $\mathcal{M}$
trained on the extended dataset $\mathcal{D}_* = \mathcal{D} \cup (x, y)$.
For Bayesian linear regression, the posterior entropy and the predictive variance
are commonly used as measures of the model variance---both are available in closed-form
and correspond to well-known alphabetic optimal design criteria~\cite{chaloner1995bayesian}.
Under mild assumptions~\cite{cohn1996active},
closed-form expressions for the output variance can also be obtained
for more flexible models, such as neural networks, Gaussian mixture models,
and locally-weighted linear regression~\cite{settles2009active}.

\subsection{Bayesian Linear Regression}
\label{subsec:regression}
Linear regression~\cite{bishop2006pattern} assumes
that the output $y \in \mathbb{R}$ is given by a linear function $\cvec{\theta}^T\cvec{\phi}(\cvec{x})$
in the features $\cvec{\phi}(\cvec{x}) \in \mathbb{R}^m$ of the input $\cvec{x} \in \mathbb{R}^n$
and parameters $\cvec{\theta} \in \mathbb{R}^m$.
The output uncertainty is captured by a probability distribution, most commonly---the normal distribution,
\begin{align*}
    p(y \given \cvec{x}; \cvec{\theta}) =
        \mathcal{N}\left(y \given \cvec{\theta}^T\cvec{\phi}(\cvec{x}), \beta^{-1} \right).
\end{align*}
In Bayesian linear regression~\cite{bishop2006pattern},
not only the output $y$ but also the parameters $\cvec{\theta}$ are assumed to be uncertain.
Conveniently, given a Gaussian prior
$p(\cvec{\theta} \given \mathcal{D}) =
    \mathcal{N}\left(\cvec{\theta} \given \cvec{\mu}, \cmat{\Sigma}\right)$,
the \emph{posterior} after observing a data point $(\cvec{x}, y)$ is also Gaussian,
$p(\cvec{\theta} \given \mathcal{D} \cup (\cvec{x}, y)) =
\mathcal{N}\left(\cvec{\theta} \given \cvec{\mu}_*, \cmat{\Sigma}_* \right)$, with parameters
\begin{equation}
    \label{eq:blrcov}
    \begin{aligned}
        \cvec{\mu}_* &=
        \cmat{\Sigma}_* \left(\cmat{\Sigma}^{-1} \cvec{\mu} + \beta \cmat{\Phi}^T \cmat{Y}\right),\\
    \cmat{\Sigma}_*^{-1} &=
        \cmat{\Sigma}^{-1} + \beta \cmat{\Phi}^T \cmat{\Phi},
    \end{aligned}
\end{equation}
where the features and targets corresponding to the new data point $(\cvec{x}, y)$
are aggregated into the design matrix $\cvec{\Phi}$ and the vector of targets $\cmat{Y}$,
as described in~\cite{bishop2006pattern}.
Along with the posterior, the \emph{predictive distribution} plays an important role,
\begin{align*}
    p(y_* \given \cvec{x}_*; \mathcal{D} \cup (\cvec{x}, y)) &=
        \mathcal{N}(y_* \given \cvec{\mu}_*^T\cvec{\phi}(\cvec{x}_*), \sigma^2_*(\cvec{x}_*)),\\
    \mathrm{where\ } \quad \sigma^2_*(\cvec{x}_*) &=
        \beta^{-1} + \cvec{\phi}(\cvec{x}_*)^T \cmat{\Sigma}_* \cvec{\phi}(\cvec{x}_*).
\end{align*}
If data arrives sequentially, the posterior can be updated incrementally,
taking the current posterior as the prior for the next data point.
This procedure, known as iterative least squares~\cite{bishop2006pattern},
enables efficient processing of sequential data.

\subsection{Random Fourier Features}
\label{subsec:rff}
Random Fourier features provide a powerful representation by approximating
a Gaussian process with an exponentiated quadratic kernel~\cite{rahimi2008random}.
Given an input $\cvec{x} \in \mathbb{R}^n$, following~\cite{rajeswaran2017towards},
the $i$-th feature is defined as
$\phi_i(\cvec{x}) = \sin (\sum^n_{j=1} P_{i j} x_j/\nu_j + \varphi_i)$
where $P_{i j} \sim \mathcal{N}(0,1)$ are normally distributed real numbers
and $\varphi_i \sim \mathcal{U}[-\pi, \pi)$ are uniform random phase shifts.
The bandwidth parameter $\cvec{\nu} \in \mathbb{R}^n$ scales the inputs,
and must be chosen with care.
A rule of thumb suggested in~\cite{rajeswaran2017towards}
is to set it equal to the average pairwise distance between observed input vectors.
An even better approach is to fit the bandwidth parameter through
marginal maximum likelihood optimization.

\subsection{Model-Based Reinforcement Learning}
The distinction between model-free and model-based reinforcement learning
is not precisely defined, but for the purposes of this paper,
under model-based reinforcement learning we will understand
trajectory optimization using a dynamics model obtained through system identification.
In detail, having learned a forward model $s' = f_\theta(s, a)$,
the agent plans a trajectory $\tau = \left\{s_0, a_0, s_1, a_1, \dots, a_{T-1}, s_T\right\}$
from a given starting state $s_0$ with the goal of minimizing a pre-defined trajectory cost $J(\tau)$.
Variations are possible in how exactly the model is learned,
what model representation is used,
whether the cost function is learned or given, etc.

\subsection{Shooting Methods for Trajectory Optimization}
\label{subsec:shooting}
Direct shooting methods are a class of optimal control algorithms aimed at solving
planning problems with deterministic continuous dynamics models~\cite{stoer2013introduction}.
Depending on whether the dynamics are imposed as a constraint or included in the objective itself,
one discerns between single shooting and multiple shooting methods.
In \emph{single shooting}, the dynamics are included in the objective function,
\begin{equation*}
    \underset{a_{0:T-1}}{\textrm{minimize}} \quad
    J(s_0, a_0, f(s_0, a_0), a_1,  \dots, a_{T-1}, f(f(\dots f(s_0, a_0), a_{T-2}), a_{T-1})),
\end{equation*}
and the optimization is performed only with respect to the actions~$a_{0:T-1}$.
Note the iterated application of the dynamics function~$f$.
When performing gradient descent on this objective, the problem
\clearpage
known as exploding/vanishing gradients hinders efficient first-order optimization.
The reason for that is the ill conditioning of the problem:
the actions in the beginning of the trajectory
have a bigger impact on the final state than the actions at the end.

\emph{Multiple shooting} methods aim to improve the problem conditioning
by splitting the trajectory into smaller chunks, subsequently glueing them together by constraints.
To enable that, the dynamics are imposed as a constraint,
and the optimization is performed with respect to both actions and states,
\begin{align*}
    \underset{a_{0:T-1}, s_{1:T}}{\textrm{minimize}} &\quad
        J(s_0, a_0, s_1, a_1,  \dots, a_{T-1}, s_{T})\\
    \textrm{subject to} &\quad
        s_t = f(s_{t-1}, a_{t-1}),\quad t=1,\dots,T.
\end{align*}
Multiple shooting methods converge faster and are numerically more stable than single shooting
because they are better conditioned thanks to the
breaking up of long-term dependencies into shorter chunks~\cite{andersson2018casadi}.
Moreover, state constraints can be straightforwardly incorporated in multiple shooting
since states are optimization variables.
The price for such benefits is a significant increase in the problem size
and, as a consequence, in the memory requirements.
This drawback, however, is offset by the fact that the problem becomes much sparser~\cite{andersson2018casadi}.

\section{Receding Horizon Curiosity}
\label{sec:rhc}

Consider a dynamical system with state space $\mathcal{S} \subset \mathbb{R}^n$
and action space $\mathcal{A} \subset \mathbb{R}^k$.
Denote a probabilistic model of the dynamics
that an agent maintains about this system by~$\mathcal{M}$.
The probabilistic nature of the model enables the agent
to reason about the information content in observations.
The agent wants to find a sequence of actions $a_{0:T-1}$
which, when executed open-loop on the real system,
provides the most informative sequence
of observations $s_{0:T}$ in the sense of being useful for learning the model.
This setting is an active learning problem
in which the sequence of actions plays the role of a query point.
Therefore, we can utilize active learning approaches from Sec.~\ref{subsec:active-learning}
based on uncertainty sampling and expected variance reduction to solve the exploration problem.

\subsection{Uncertainty Sampling}
\label{subsec:us}
The uncertainty sampling acquisition function proposes to query
the point that the model is most uncertain about.
In our planning scenario, that means selecting the sequence of actions $a_{0:T-1}$
that results in a trajectory $(a_{0:T-1}, s_{0:T})$ that has the highest entropy.
If $p(s' \given s, a)$ is the prediction given by the probabilistic model $\mathcal{M}$
trained on previously observed trajectories $\mathcal{D}$,
then the optimization objective can be stated as
\begin{equation*}
    \underset{a_{0:T-1}}{\textrm{maximize}} \quad
        \sum_{t = 1:T} \mathbb{E}_{s_{t-1} \sim p(s_{t-1} \given a_{t-2}, \dotsc, a_0)}
            \mathbb{V}_p[s_{t} \given s_{t-1}, a_{t-1}],
\end{equation*}
with $\mathbb{V}_p[s_{t} \given s_{t-1}, a_{t-1}]$ denoting the variance
in the prediction of the next state,
and the outer expectation being taken with respect to state marginal distributions.
Note that alternative formulations are possible,
e.g., where only the variance at the last time step is taken into account
or where different time steps are weighted differently.
Ideally, one would only consider the variance at the last time step;
however, due to the fact that an approximate model is used,
real trajectories rather quickly diverge from the planned ones,
and therefore it is desirable to reach informative states as quickly as possible,
which can be achieved by rewarding the agent for information gain at every time step.

To evaluate the uncertainty sampling objective, the state distribution needs to be propagated
through the probabilistic model, which is a non-trivial problem in general.
Instead, an approximate version of this problem can be solved,
where only the mean of the state distribution is propagated,
\begin{equation}
    \begin{aligned}
        \underset{a_{0:T-1}}{\textrm{maximize}} &\quad
          \sum_{t = 1:T} \mathbb{V}_p[s_{t} \given \hat{s}_{t-1}, a_{t-1}]
                \label{eq:ex-us}\\
        \textrm{subject to} &\quad
            \hat{s}_{t} = \mathbb{E}_p [s_{t} \given \hat{s}_{t-1}, a_{t-1}], \quad t=1,\dots,T.
    \end{aligned}
\end{equation}
In line with the general theory of active learning, the uncertainty sampling objective~\eqref{eq:ex-us}
is easier to optimize than the expected variance reduction objective described below.
If the Bayesian linear regression model (Sec.~\ref{subsec:regression}) is used
to represent the dynamics, the posterior parameter covariance matrix $\cmat{\Sigma}_*$
remains constant and does not depend on the states and actions.
Thus, both the objective and constraints in~\eqref{eq:ex-us} are differentiable
and the problem can be solved using the multiple shooting method
described in Sec.~\ref{subsec:shooting}.
The complete optimization procedure is summarized in Alg.~\ref{alg:sysid}.

\subsection{Expected Variance Reduction}
\label{subsec:evr}
The uncertainty sampling heuristic rewards the agent for visiting uncertain states,
but it ignores the fact that the model will become more certain once those states are visited.
For example, if two states are initially equally uncertain,
visiting one of them may yield a larger decrease in uncertainty
because that state is more informative.
Uncertainty sampling would be insensitive to this difference,
whereas expected variance reduction allows for taking such information gain into account.
In our setting, the expected variance reduction problem can be stated as
\begin{equation*}
    \underset{a_{0:T-1}}{\textrm{minimize}} \quad
        \mathbb{E}_p \left[
            \textrm{var}(\mathcal{M} \given \mathcal{D} \cup (a_{0:T-1}, s_{0:T}))
        \right].
\end{equation*}
The operator $\textrm{var}(\mathcal{M} \given \mathcal{D}_*)$ here
stands for a measure of variance of model $\mathcal{M}$ trained on dataset $\mathcal{D}_*$.
We take it to be the entropy of the posterior distribution
over the model parameters $\cvec{\theta} \in \mathbb{R}^m$,
which is known as the $D$-optimality criterion
in Bayesian experimental design~\cite{chaloner1995bayesian},
\begin{equation*}
    \textrm{var}(\mathcal{M \given \mathcal{D}_*})
        = \mathbb{H}(\cvec{\theta} \given \mathcal{D}_*)
        = \frac{1}{2} \ln \textrm{det}\,(\cmat{\Sigma}_*) + \frac{m}{2} \ln (2 \pi e).
\end{equation*}
Importantly, the covariance matrix $\cmat{\Sigma}_*$
depends on the trajectory $(a_{0:T-1}, s_{0:T})$
through the augmented dataset $\mathcal{D}_* = \mathcal{D} \cup (a_{0:T-1}, s_{0:T})$.
The exact relationship is given in~\eqref{eq:blrcov},
and this relationship allows for optimization of the expected model variance
with respect to the planned trajectory.

Although superior to uncertainty sampling in information-theoretic terms,
expected variance reduction is quite expensive to compute
and optimize in practice~\cite{settles2009active},
particularly due to probabilistic state propagation~\cite{deisenroth2011pilco}.
We adopt the so called maximum likelihood observations assumption~\cite{platt2010belief},
which amounts to propagating only the mean of the state distribution.
The crudeness of this approximation is offset by the computational advantage:
more frequent replanning, enabled by neglecting the expensive state uncertainty propagation,
allows the agent to compensate for unforeseen deviations from the planned trajectory efficiently.
The corresponding optimization problem reads
\begin{equation}
    \begin{aligned}
        \underset{a_{0:T-1}}{\textrm{minimize}} &\quad
            \textrm{var}(\mathcal{M} \given \mathcal{D} \cup (a_{0:T-1}, s_{0:T}))
            \label{eq:ex-mvr}\\
        \textrm{subject to} &\quad
            \hat{s}_{t} = \mathbb{E}_p [s_{t} \given \hat{s}_{t-1}, a_{t-1}], \quad t=1,\dots,T.
    \end{aligned}
\end{equation}
Both the objective and constraints in~\eqref{eq:ex-mvr} are differentiable.
Therefore, gradient-based optimization described in Sec.~\ref{subsec:shooting}
can in principle be used to solve this problem.
However, evaluation of the objective requires differentiation through matrix inversion,
since matrix $\cmat{\Sigma}_*$ depends on the inverse of the kernel matrix~\eqref{eq:blrcov}.
Combined with the chain-like structure of state-action dynamics,
gradient computation becomes quite expensive for larger models (e.\,g.\ $T > 100$, $M > 40$).

\begin{algorithm}[t]
    \KwData{number of episodes $N$, horizon $T$, initial model $\mathcal{M}_0$}
    \KwResult{optimized model $\mathcal{M}_N$}
    \For{$i\gets1$ \KwTo $N$}{
        find actions $a_{0:T-1}$ that optimize~\eqref{eq:ex-us}
            or~\eqref{eq:ex-mvr} given the current model $\mathcal{M}_i$\;
        execute $a_{0:T-1}$ in the environment and observe $s_{0:T}$\;
        update model $\mathcal{M}_{i+1}$ via~\eqref{eq:blrcov}
            using $\mathcal{M}_i$ as the prior
            and $(a_{0:T-1}, s_{0:T})$ as the new data\;
    }
    \caption{Receding Horizon Curiosity.
        In each episode $i$, the most informative sequence of actions $a_{0:T-1}$
        under the current model $\mathcal{M}_i$ is computed;
        after that, observations $s_{0:T}$ are collected
        and the model is updated $\mathcal{M}_i \to \mathcal{M}_{i+1}$.
    }
    \label{alg:sysid}
\end{algorithm}

\section{Experimental Results}
\label{sec:result}

We compare our receding horizon curiosity algorithm (RHC, Sec.~\ref{sec:rhc})
to state-of-the-art model-based and model-free exploration approaches.
On the model-based side, we consider MAX~\cite{shyam2018model},
which optimizes a certain approximation
of the information gain via model ensemble disagreement.
On the model-free side, we employ soft actor-critic (SAC)~\cite{haarnoja2018soft}
with popular exploration bonuses:
squared prediction error (SAC PE) and information gain in the form of
parameter entropy difference between successive steps (SAC IG).
Two acquisition functions for RHC are considered:
\emph{uncertainty sampling} (RHC US, Sec.~\ref{subsec:us}) and
\emph{expected variance reduction} (RHC EVR, Sec.~\ref{subsec:evr}).
Additionally, we report the performance of
uniform random exploration (RAND) for comparison.
\clearpage
\newlength\halfheight
\newlength\halfwidth
\setlength\halfheight{6cm}
\setlength\halfwidth{.5\linewidth}
\begin{figure}[t]
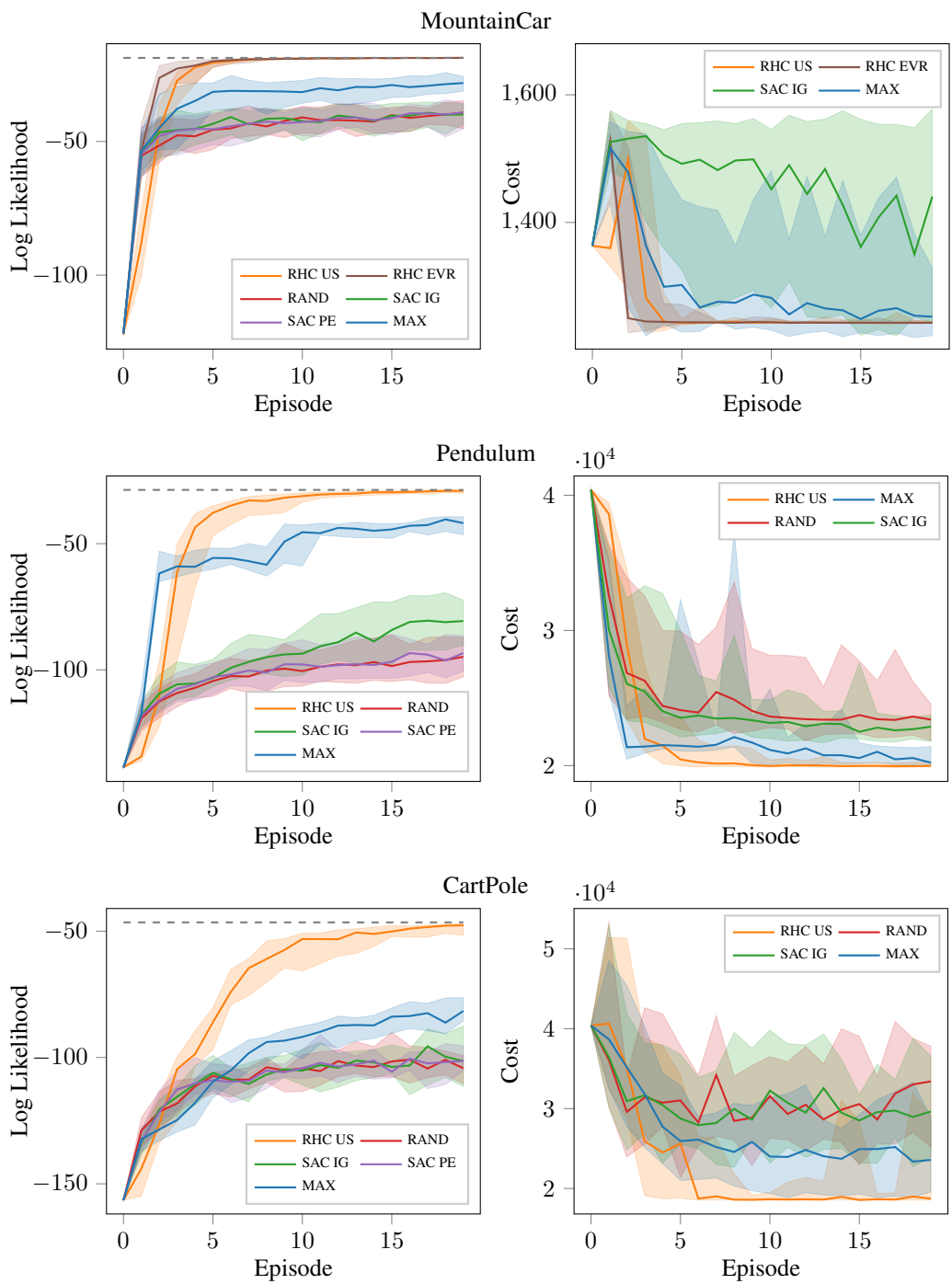

	\begin{center}
		\vspace{20pt}
		MountainCar\\
	\end{center}
	\input{img/mc_nstep_llh.tex}
	\input{img/mc_cost.tex}
	\begin{center}
		Pendulum\\
		\vspace{-10pt}
	\end{center}
	\input{img/pend_nstep_llh.tex}
\begin{tikzpicture}

\definecolor{col_max}{rgb}{0.12156862745098,0.466666666666667,0.705882352941177}  
\definecolor{col_us}{rgb}{1,0.498039215686275,0.0549019607843137}  
\definecolor{col_ig}{rgb}{0.172549019607843,0.627450980392157,0.172549019607843}  
\definecolor{col_rand}{rgb}{0.83921568627451,0.152941176470588,0.156862745098039}  
\definecolor{col_pe}{rgb}{0.580392156862745,0.403921568627451,0.741176470588235}  
\definecolor{col_vr}{rgb}{0.549019607843137,0.337254901960784,0.294117647058824}  

\begin{axis}[
height=\halfheight,
legend cell align={left},
legend entries={{RHC US},{MAX},{RAND},{SAC IG}},
legend style={draw=white!80.0!black, font=\tiny, legend columns=2, line width=.8pt},
tick align=outside,
tick pos=left,
width=\halfwidth,
x grid style={white!69.01960784313725!black},
xlabel={Episode},
xmin=-0.95, xmax=19.95,
y grid style={white!69.01960784313725!black},
ylabel={Cost},
ylabel style={yshift=-3mm},
ymin=18826.1074631424, ymax=41427.3282160408
]
\addlegendimage{no markers, col_us}
\addlegendimage{no markers, col_max}
\addlegendimage{no markers, col_rand}
\addlegendimage{no markers, col_ig}
\path [draw=col_us, fill=col_us, opacity=0.2] (axis cs:0,40400)
--(axis cs:0,40400)
--(axis cs:1,34703.9130069911)
--(axis cs:2,23418.1729849161)
--(axis cs:3,21040.0456262495)
--(axis cs:4,20149.2715721376)
--(axis cs:5,19972.8753730602)
--(axis cs:6,19895.0942856981)
--(axis cs:7,19957.9627697309)
--(axis cs:8,19940.7143884972)
--(axis cs:9,19901.8333030689)
--(axis cs:10,19889.4276558708)
--(axis cs:11,19904.797059741)
--(axis cs:12,19899.9868594315)
--(axis cs:13,19876.0728580956)
--(axis cs:14,19894.0940251102)
--(axis cs:15,19891.022526639)
--(axis cs:16,19891.9925266338)
--(axis cs:17,19853.4356791832)
--(axis cs:18,19888.7028480265)
--(axis cs:19,19893.9011477312)
--(axis cs:19,20321.3338691413)
--(axis cs:19,20321.3338691413)
--(axis cs:18,20246.8147692735)
--(axis cs:17,20171.6636562447)
--(axis cs:16,20235.7059997632)
--(axis cs:15,20181.1600995613)
--(axis cs:14,20241.1137365929)
--(axis cs:13,20674.8203456574)
--(axis cs:12,20373.4664658999)
--(axis cs:11,20538.3641413435)
--(axis cs:10,20407.6613167853)
--(axis cs:9,21258.8195805862)
--(axis cs:8,21472.299479067)
--(axis cs:7,21358.8196272682)
--(axis cs:6,21903.8815940398)
--(axis cs:5,23639.3523934406)
--(axis cs:4,23571.3137613445)
--(axis cs:3,27396.4591416855)
--(axis cs:2,34595.1263510483)
--(axis cs:1,39452.6806665976)
--(axis cs:0,40400)
--cycle;

\path [draw=col_max, fill=col_max, opacity=0.2] (axis cs:0,40400)
--(axis cs:0,40400)
--(axis cs:1,25387.2591311526)
--(axis cs:2,20466.3287805945)
--(axis cs:3,20909.8522643741)
--(axis cs:4,21239.6774928228)
--(axis cs:5,21077.5838607671)
--(axis cs:6,21056.6446041849)
--(axis cs:7,21145.9442738504)
--(axis cs:8,21374.3129288582)
--(axis cs:9,20191.6972023028)
--(axis cs:10,20058.0607779036)
--(axis cs:11,20215.6772993115)
--(axis cs:12,19988.0802048834)
--(axis cs:13,20174.8532168213)
--(axis cs:14,20106.8962882793)
--(axis cs:15,20154.7409008786)
--(axis cs:16,20210.0739874878)
--(axis cs:17,20099.2273937319)
--(axis cs:18,20027.7817342932)
--(axis cs:19,19983.2545873513)
--(axis cs:19,21398.9086253309)
--(axis cs:19,21398.9086253309)
--(axis cs:18,21324.4186141354)
--(axis cs:17,21369.8152721772)
--(axis cs:16,21442.3281336663)
--(axis cs:15,21682.7216130659)
--(axis cs:14,24210.9506957154)
--(axis cs:13,22876.198559548)
--(axis cs:12,23088.7648665377)
--(axis cs:11,22107.8123041628)
--(axis cs:10,25650.4034594793)
--(axis cs:9,23521.9455679125)
--(axis cs:8,37371.0636266723)
--(axis cs:7,22499.4965951655)
--(axis cs:6,27660.9895968331)
--(axis cs:5,32273.0564989464)
--(axis cs:4,24409.0344236865)
--(axis cs:3,25729.67378716)
--(axis cs:2,24421.0600978362)
--(axis cs:1,35324.8690435998)
--(axis cs:0,40400)
--cycle;

\path [draw=col_rand, fill=col_rand, opacity=0.2] (axis cs:0,40400)
--(axis cs:0,40400)
--(axis cs:1,25154.7921646154)
--(axis cs:2,23429.1649417028)
--(axis cs:3,23659.7647166168)
--(axis cs:4,23219.9782010279)
--(axis cs:5,22687.1658362864)
--(axis cs:6,22486.1572687321)
--(axis cs:7,23010.7114998722)
--(axis cs:8,22368.356841935)
--(axis cs:9,22554.7178247756)
--(axis cs:10,22800.8762598307)
--(axis cs:11,22696.1335067213)
--(axis cs:12,21826.080341831)
--(axis cs:13,22289.9558743663)
--(axis cs:14,22352.247683628)
--(axis cs:15,22357.0869410689)
--(axis cs:16,22057.6157732343)
--(axis cs:17,22112.3966113972)
--(axis cs:18,22250.351802505)
--(axis cs:19,21843.8165560543)
--(axis cs:19,24556.4714940767)
--(axis cs:19,24556.4714940767)
--(axis cs:18,26211.024402759)
--(axis cs:17,28593.0633008653)
--(axis cs:16,26293.7136967547)
--(axis cs:15,27737.5216198568)
--(axis cs:14,28935.3314404591)
--(axis cs:13,25820.2577927346)
--(axis cs:12,27997.2703557719)
--(axis cs:11,28143.5438922513)
--(axis cs:10,28156.4659816181)
--(axis cs:9,28659.1641103064)
--(axis cs:8,33572.1738794122)
--(axis cs:7,30249.1065293012)
--(axis cs:6,28966.2866508681)
--(axis cs:5,29939.121422946)
--(axis cs:4,30005.417021448)
--(axis cs:3,32546.2807621585)
--(axis cs:2,33922.2141123207)
--(axis cs:1,36355.6993602499)
--(axis cs:0,40400)
--cycle;

\path [draw=col_ig, fill=col_ig, opacity=0.2] (axis cs:0,40400)
--(axis cs:0,40400)
--(axis cs:1,25780.1851627859)
--(axis cs:2,23328.1672397082)
--(axis cs:3,23184.8564537679)
--(axis cs:4,22698.4512621366)
--(axis cs:5,22114.7688549695)
--(axis cs:6,22246.9260257894)
--(axis cs:7,22357.2341657901)
--(axis cs:8,22636.5510950533)
--(axis cs:9,22111.8062339282)
--(axis cs:10,21980.0035612895)
--(axis cs:11,21806.1645139993)
--(axis cs:12,21880.4720491846)
--(axis cs:13,21675.326273243)
--(axis cs:14,22041.9179744027)
--(axis cs:15,21979.7298142778)
--(axis cs:16,21788.3998296275)
--(axis cs:17,21745.6585810483)
--(axis cs:18,21967.197970933)
--(axis cs:19,21775.6813241205)
--(axis cs:19,23747.5163153371)
--(axis cs:19,23747.5163153371)
--(axis cs:18,23573.727345811)
--(axis cs:17,23711.6953269899)
--(axis cs:16,23864.6838777137)
--(axis cs:15,27588.6345617439)
--(axis cs:14,24033.617368302)
--(axis cs:13,24006.1945262655)
--(axis cs:12,25208.8291303428)
--(axis cs:11,25609.8308405587)
--(axis cs:10,24863.540843524)
--(axis cs:9,24890.6984896596)
--(axis cs:8,29619.2475472926)
--(axis cs:7,26260.5012393403)
--(axis cs:6,26681.710983265)
--(axis cs:5,30010.1005745112)
--(axis cs:4,32744.9643490449)
--(axis cs:3,33301.0126577689)
--(axis cs:2,32393.3122243716)
--(axis cs:1,36083.6940926007)
--(axis cs:0,40400)
--cycle;

\addplot [thick, col_us]
table [row sep=\\]{%
0	40400 \\
1	38616.7610505997 \\
2	28983.4861978183 \\
3	21979.0116935716 \\
4	21474.9719046312 \\
5	20450.7341536742 \\
6	20236.9354092998 \\
7	20146.2940220444 \\
8	20157.0609433157 \\
9	20032.8241830735 \\
10	19965.9894275555 \\
11	20020.4862928133 \\
12	20020.4256563731 \\
13	20007.4798724278 \\
14	19958.9626186271 \\
15	19974.1078106283 \\
16	19975.5928533612 \\
17	19961.4436218262 \\
18	19959.8947730947 \\
19	19975.9737897236 \\
};
\addplot [thick, col_max]
table [row sep=\\]{%
0	40400 \\
1	28120.808497745 \\
2	21358.7795858417 \\
3	21406.929197959 \\
4	21510.7474980834 \\
5	21468.4835325356 \\
6	21393.5089908419 \\
7	21536.7501756062 \\
8	22104.4924987903 \\
9	21703.5443450292 \\
10	21158.3004369996 \\
11	20898.2516982039 \\
12	21271.3296334589 \\
13	20763.4214253879 \\
14	20763.3649274283 \\
15	20552.6795278118 \\
16	21018.6907477457 \\
17	20463.7344771578 \\
18	20557.4431976772 \\
19	20212.4108451331 \\
};
\addplot [thick, col_rand]
table [row sep=\\]{%
0	40400 \\
1	32561.435199687 \\
2	26852.8735150705 \\
3	26274.1953255208 \\
4	24413.534220123 \\
5	24102.3677134741 \\
6	23918.6531992653 \\
7	25439.2907754506 \\
8	24884.724286054 \\
9	24047.4308409288 \\
10	23629.0480501075 \\
11	23522.8363856846 \\
12	23430.4958874191 \\
13	23392.2122275529 \\
14	23399.0512335677 \\
15	23738.7430511434 \\
16	23431.7963046247 \\
17	23377.8034386898 \\
18	23627.7924501341 \\
19	23397.9975597145 \\
};
\addplot [thick, col_ig]
table [row sep=\\]{%
0	40400 \\
1	30044.0316457291 \\
2	26055.0878095435 \\
3	25472.3443408815 \\
4	24021.727867867 \\
5	23529.6278659403 \\
6	23699.2651350084 \\
7	23476.6400050973 \\
8	23513.3379166109 \\
9	23352.4850893744 \\
10	23152.2860761562 \\
11	23226.7831716537 \\
12	22909.7534219734 \\
13	23099.0684952358 \\
14	23067.5591009784 \\
15	22498.2341048168 \\
16	22807.4533630559 \\
17	22605.7487823163 \\
18	22693.2146671437 \\
19	22877.9966358629 \\
};

\end{axis}

\end{tikzpicture}
	\begin{center}
		CartPole\\
		\vspace{-10pt}
	\end{center}
	\input{img/cp_nstep_llh.tex}
\begin{tikzpicture}

\definecolor{col_max}{rgb}{0.12156862745098,0.466666666666667,0.705882352941177}  
\definecolor{col_us}{rgb}{1,0.498039215686275,0.0549019607843137}  
\definecolor{col_ig}{rgb}{0.172549019607843,0.627450980392157,0.172549019607843}  
\definecolor{col_rand}{rgb}{0.83921568627451,0.152941176470588,0.156862745098039}  
\definecolor{col_pe}{rgb}{0.580392156862745,0.403921568627451,0.741176470588235}  
\definecolor{col_vr}{rgb}{0.549019607843137,0.337254901960784,0.294117647058824}  

\begin{axis}[
height=\halfheight,
legend cell align={left},
legend entries={{RHC US},{RAND},{SAC IG},{MAX}},
legend style={draw=white!80.0!black, font=\tiny, legend columns=2, line width=.8pt},
tick align=outside,
tick pos=left,
width=\halfwidth,
x grid style={white!69.01960784313725!black},
xlabel={Episode},
xmin=-0.95, xmax=19.95,
y grid style={white!69.01960784313725!black},
ylabel={Cost},
ylabel style={yshift=-3mm},
ymin=16805.6713984519, ymax=55049.4565998772
]
\addlegendimage{no markers, col_us}
\addlegendimage{no markers, col_rand}
\addlegendimage{no markers, col_ig}
\addlegendimage{no markers, col_max}
\path [draw=col_us, fill=col_us, opacity=0.2] (axis cs:0,40400)
--(axis cs:0,40400)
--(axis cs:1,32397.0539936043)
--(axis cs:2,28741.1361917838)
--(axis cs:3,19092.5982091601)
--(axis cs:4,18700.7164513047)
--(axis cs:5,18779.016001728)
--(axis cs:6,18550.207181042)
--(axis cs:7,18573.386161075)
--(axis cs:8,18550.0381513251)
--(axis cs:9,18547.5114353994)
--(axis cs:10,18546.7141875744)
--(axis cs:11,18545.5425608034)
--(axis cs:12,18545.0655332673)
--(axis cs:13,18548.2700571233)
--(axis cs:14,18583.8514363032)
--(axis cs:15,18544.3124105312)
--(axis cs:16,18544.852741136)
--(axis cs:17,18544.025271244)
--(axis cs:18,18546.6246935413)
--(axis cs:19,18545.6183670604)
--(axis cs:19,29060.5024257918)
--(axis cs:19,29060.5024257918)
--(axis cs:18,31952.8594687774)
--(axis cs:17,19455.3799387266)
--(axis cs:16,19380.8309715772)
--(axis cs:15,28464.6862506103)
--(axis cs:14,30989.8503164231)
--(axis cs:13,20905.3211049718)
--(axis cs:12,21405.8968197562)
--(axis cs:11,20792.0901570287)
--(axis cs:10,19484.7263334412)
--(axis cs:9,19158.7606686204)
--(axis cs:8,21849.72963921)
--(axis cs:7,26378.6172775319)
--(axis cs:6,26385.0790190392)
--(axis cs:5,34229.8947129737)
--(axis cs:4,37820.6805098302)
--(axis cs:3,38871.0047016101)
--(axis cs:2,51335.4235201556)
--(axis cs:1,51415.3664794358)
--(axis cs:0,40400)
--cycle;

\path [draw=col_rand, fill=col_rand, opacity=0.2] (axis cs:0,40400)
--(axis cs:0,40400)
--(axis cs:1,30223.9319344171)
--(axis cs:2,23921.3731751995)
--(axis cs:3,25581.233946245)
--(axis cs:4,27735.9213408711)
--(axis cs:5,21409.5564053667)
--(axis cs:6,25339.6198571799)
--(axis cs:7,26562.9844331022)
--(axis cs:8,26007.6450273578)
--(axis cs:9,22875.057389436)
--(axis cs:10,26282.6102889517)
--(axis cs:11,26085.6941036495)
--(axis cs:12,24735.6332833243)
--(axis cs:13,26144.8815839034)
--(axis cs:14,25818.7268442255)
--(axis cs:15,22936.9351813966)
--(axis cs:16,25578.3338195628)
--(axis cs:17,25557.6559737983)
--(axis cs:18,27036.0035303553)
--(axis cs:19,25222.0544227708)
--(axis cs:19,37811.1249669067)
--(axis cs:19,37811.1249669067)
--(axis cs:18,38780.6429755323)
--(axis cs:17,40905.5280221435)
--(axis cs:16,34002.6532450949)
--(axis cs:15,39073.62149535)
--(axis cs:14,39975.9344596261)
--(axis cs:13,35623.4025736655)
--(axis cs:12,37761.9171030457)
--(axis cs:11,35853.9532870662)
--(axis cs:10,36505.2457037404)
--(axis cs:9,35386.1550332113)
--(axis cs:8,34063.0849257251)
--(axis cs:7,41619.2291864335)
--(axis cs:6,34175.9342760907)
--(axis cs:5,37879.2981695648)
--(axis cs:4,41850.7729765633)
--(axis cs:3,42629.6575209277)
--(axis cs:2,34163.1351434191)
--(axis cs:1,53311.1027270851)
--(axis cs:0,40400)
--cycle;

\path [draw=col_ig, fill=col_ig, opacity=0.2] (axis cs:0,40400)
--(axis cs:0,40400)
--(axis cs:1,30172.4235181579)
--(axis cs:2,24836.5128673862)
--(axis cs:3,26864.371592946)
--(axis cs:4,23185.6118151831)
--(axis cs:5,25536.4194488821)
--(axis cs:6,24978.867232813)
--(axis cs:7,23041.6546859921)
--(axis cs:8,22653.3260863866)
--(axis cs:9,26273.8188073284)
--(axis cs:10,23151.6817422621)
--(axis cs:11,25146.4055688743)
--(axis cs:12,23916.0939555435)
--(axis cs:13,24859.2086024419)
--(axis cs:14,25022.0150540136)
--(axis cs:15,23231.4132393589)
--(axis cs:16,23129.7616594333)
--(axis cs:17,26360.9353014349)
--(axis cs:18,22723.0338677801)
--(axis cs:19,23955.6777924577)
--(axis cs:19,36545.8664941823)
--(axis cs:19,36545.8664941823)
--(axis cs:18,38802.5360543302)
--(axis cs:17,33670.6013624795)
--(axis cs:16,35539.1067249116)
--(axis cs:15,31735.4446581708)
--(axis cs:14,36180.4433969215)
--(axis cs:13,39327.6124953823)
--(axis cs:12,37950.9701061187)
--(axis cs:11,38107.9941728581)
--(axis cs:10,39820.4968762434)
--(axis cs:9,37548.6549050851)
--(axis cs:8,39539.4961642737)
--(axis cs:7,34750.4738768286)
--(axis cs:6,33902.4375491836)
--(axis cs:5,36798.6581969823)
--(axis cs:4,37979.2952297274)
--(axis cs:3,38164.8433841102)
--(axis cs:2,41912.9372634786)
--(axis cs:1,52876.0530613277)
--(axis cs:0,40400)
--cycle;

\path [draw=col_max, fill=col_max, opacity=0.2] (axis cs:0,40400)
--(axis cs:0,40400)
--(axis cs:1,32500.2697270782)
--(axis cs:2,27779.2574316728)
--(axis cs:3,25131.6964554952)
--(axis cs:4,23100.632657688)
--(axis cs:5,21006.2523022273)
--(axis cs:6,21119.5672456829)
--(axis cs:7,21301.0595992951)
--(axis cs:8,20313.7311622691)
--(axis cs:9,19705.8849492934)
--(axis cs:10,19365.6515547872)
--(axis cs:11,18834.5530665117)
--(axis cs:12,19474.2625345869)
--(axis cs:13,19182.9814979304)
--(axis cs:14,19074.3092144409)
--(axis cs:15,19285.9392319762)
--(axis cs:16,18849.7913280468)
--(axis cs:17,19040.9126212714)
--(axis cs:18,18894.5694015931)
--(axis cs:19,19577.0039942768)
--(axis cs:19,28944.8333715079)
--(axis cs:19,28944.8333715079)
--(axis cs:18,33303.0538519749)
--(axis cs:17,32114.7976912323)
--(axis cs:16,31113.0561310122)
--(axis cs:15,29255.0472429263)
--(axis cs:14,27524.5649940779)
--(axis cs:13,28595.8963200588)
--(axis cs:12,33002.5214650164)
--(axis cs:11,31841.1871851074)
--(axis cs:10,31404.3470433513)
--(axis cs:9,30395.0345623577)
--(axis cs:8,30653.9503046736)
--(axis cs:7,29459.3204204966)
--(axis cs:6,29039.3064436862)
--(axis cs:5,32883.5180424753)
--(axis cs:4,34460.9818566765)
--(axis cs:3,40275.0543534111)
--(axis cs:2,45456.3081907595)
--(axis cs:1,48497.635683492)
--(axis cs:0,40400)
--cycle;

\addplot [thick, col_us]
table [row sep=\\]{%
0	40400 \\
1	40634.3491947482 \\
2	35182.7929446991 \\
3	25902.5390292142 \\
4	24510.9903500932 \\
5	25633.7293838085 \\
6	18736.9811038296 \\
7	19008.2731894769 \\
8	18601.4342393499 \\
9	18580.9396258311 \\
10	18643.6520478441 \\
11	18616.4785025765 \\
12	18636.3783180675 \\
13	18607.7682157424 \\
14	18949.6594897527 \\
15	18558.1576478755 \\
16	18648.1331684582 \\
17	18610.6320503665 \\
18	18975.4601323917 \\
19	18729.5338081678 \\
};
\addplot [thick, col_rand]
table [row sep=\\]{%
0	40400 \\
1	36082.5593898946 \\
2	29587.128732899 \\
3	31410.4263272553 \\
4	30738.1317881607 \\
5	31015.9230532985 \\
6	28254.0815648165 \\
7	34184.8889729167 \\
8	28453.5605864205 \\
9	28846.8699654556 \\
10	31554.290438315 \\
11	29324.865005035 \\
12	30482.4425561007 \\
13	28643.9805355692 \\
14	29854.9775593501 \\
15	30566.5467201801 \\
16	28635.9802226305 \\
17	31874.2207865411 \\
18	33022.761464676 \\
19	33410.7359345857 \\
};
\addplot [thick, col_ig]
table [row sep=\\]{%
0	40400 \\
1	36443.0677285164 \\
2	30928.7328051347 \\
3	31668.6271162374 \\
4	30442.1279976943 \\
5	28753.0282992486 \\
6	27922.9106499843 \\
7	28192.3408538568 \\
8	29950.4930845484 \\
9	28614.0187231709 \\
10	32248.5998680329 \\
11	30706.0172674816 \\
12	29490.2519493151 \\
13	32556.4319034106 \\
14	29484.2161238609 \\
15	28514.8499878693 \\
16	29561.0919765052 \\
17	29746.4218209816 \\
18	28950.4841151476 \\
19	29641.1469870727 \\
};
\addplot [thick, col_max]
table [row sep=\\]{%
0	40400 \\
1	38668.0349669165 \\
2	35176.9026633924 \\
3	31810.4787137296 \\
4	27742.5535087746 \\
5	25927.4529743228 \\
6	26108.4206124646 \\
7	25183.6907483708 \\
8	24568.8681002294 \\
9	25826.3239823552 \\
10	23998.4810327765 \\
11	23944.6155236532 \\
12	24820.4567049017 \\
13	24037.7864600845 \\
14	23710.2832380372 \\
15	24927.7309510687 \\
16	24932.927169407 \\
17	25186.4535741043 \\
18	23357.1873683886 \\
19	23574.2873808329 \\
};

\end{axis}

\end{tikzpicture}
	\caption{Evaluation of exploration methods.
		Log-likelihood of random $10$-step trajectories (evaluated by using the learned models) is shown on the left;
		the plots depict the median with the $1$st and $9$th deciles over $20$ runs.
		The grey dashed lines indicate the log-likelihood obtained
		by a model trained on $10^4$ uniform transition samples
		from the full state-action space and therefore approximates
		the best achievable log-likelihood for this model class.
		Our proposed approach RHC reaches the highest log-likelihood the fastest,
		followed by MAX, and subsequently the model-free algorithms with exploration bonuses.
		The plots on the right show the cumulative cost (negative reward) of solving each respective control task
		using the learned model.
		The trend is similar to the model log-likelihood:
		RHC reaches the lowest cost the fastest, then follows MAX,
		and after that follow the model-free exploration approaches.}
	\label{fig:eval_results}
\end{figure}
\clearpage
Experiments aim to prove the feasibility and reveal advantages and disadvantages
of simultaneous feature learning combined with approximate belief-space planning
under deterministic state propagation assumption for guiding exploration.
Classical control environments---mountain car, pendulum, and cartpole---are
used for evaluation.
In these environments, exploration using current reinforcement learning algorithms
with sparse rewards and random noise is insufficient.
\newlength\figpendheight
\newlength\figpendwidth
\setlength\figpendheight{4.84cm}
\setlength\figpendwidth{.35\linewidth}
\begin{figure}[t]
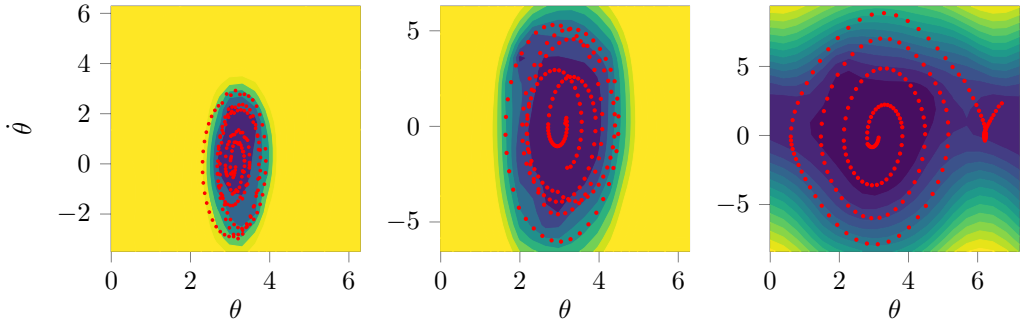

\input{img/pend2.tex}
\input{img/pend3.tex}
\input{img/pend6.tex}
\caption{Exploration progress of the RHC algorithm with the uncertainty
sampling objective in the pendulum environment.
Episodes $2$, $3$, and $6$ are shown from left to right.
Note how the explored area of the state space is growing with iterations
as the agent is trying to reach more distant states.}
\label{fig:rhc_pend}
\end{figure}
The experiments were carried out as follows.
For RHC, a sequence of actions was computed
in each episode using multiple shooting (Sec.~\ref{subsec:shooting})
and executed open-loop in the environment (Algorithm~\ref{alg:sysid}).
As planning with the expected variance reduction
objective is only tractable for small models (Sec.~\ref{subsec:evr}),
RHC with this objective was only applied to the mountain car problem.
For MAX and SAC, instead of open-loop actions,
the respective `curious' policies were applied in the environment.
After each episode,
the Bayesian linear regression model (Sec.~\ref{subsec:regression})
with random Fourier features (Sec.~\ref{subsec:rff}) was retrained
using all observations from the past episodes.
The quality of the learned model was evaluated on two metrics:
(i) \emph{mean log-likelihood} on a set of test points
obtained by sampling random starting states and executing random actions,
and (ii) \emph{mean return} (negative reward accumulated over transitions)
on downstream learning tasks, to highlight the quality of the learned model
when used in a classical planning-for-control scenario.
Further experimental details can be found in the appendix,
and the implementation---in the accompanying software package.

The results are shown in Fig.~\ref{fig:eval_results}.
In each environment, RHC was the only algorithm that reached
the highest possible log-likelihood within $20$ episodes,
as indicated by the curves in the left column.
When applicable, RHC EVR converged faster, as expected (Sec.~\ref{subsec:evr}).
Both SAC PE and SAC IG performed on the level of random exploration (RAND),
which can be traced back to their \emph{over-commitment}
behaviour~\cite{shyam2018model}:
in the beginning, virtually any action provides high reward
because the model is uncertain, but afterwards the agent has to `unlearn' it
in order to go to more distant areas of the state-action space.
MAX does not suffer from the over-commitment problem
and therefore performed better than SAC and random exploration
but nevertheless worse than our RHC method.
Although none of the methods is real-time capable,
it is worth pointing out that MAX took significantly longer than RHC
because MAX solves an entire reinforcement learning problem in each episode.
A table with run times is provided in the appendix.

Figure~\ref{fig:rhc_pend} shows the trajectories executed by RHC US
over iterations in the pendulum environment.
The background indicates the entropy of the learned forward model's output when zero action is substituted.
Warm colors correspond to high entropy (uncertainty).
RHC tries to find a trajectory of maximum uncertainty consistent with the learned dynamics model.
Driven by curiosity, the pendulum does a full swing-up already
in Episode $6$ to reach further areas of the state space.

\section{Related Work}
\label{sec:rw}

In psychology, \emph{curiosity}~\cite{silvia2012curiosity} is considered to be a type of
\emph{intrinsic motivation}~\cite{ryan2000intrinsic} that drives humans to explore.
In reinforcement learning, various reward signals have been proposed to promote \emph{artificial curiosity}.
An early example is the~\emph{prediction error}~\cite{schmidhuber1991possibility},
the idea being to reward the agent whenever there is a mismatch between predicted and observed next states.
Unfortunately, such approach suffers from the ``noisy TV problem'':
if the environment is stochastic, the agent gets attracted to the source of noise.
A cure was proposed in~\cite{schmidhuber1991curious}, which consisted in rewarding the agent
for \emph{prediction improvement} instead of prediction error.
However, despite its theoretical appeal, prediction improvement is hard to compute in practice,
especially with general function approximators~\cite{pathak2017curiosity}.

In statistics and control engineering, the problem of `optimal' exploration
is known as \emph{optimal input design}~\cite{mehra1974optimal,gevers2006input}
or \emph{optimal experiment design}~\cite{zarrop1979optimal,huan2016sequential}.
A popular measure of novelty in these fields is the \emph{information gain}~\cite{lindley1956measure}.
In computer science, the problem of `optimal' exploration is addressed by
\emph{Bayesian reinforcement learning}~\cite{ghavamzadeh2015bayesian}.
The general \emph{dual control} solution~\cite{feldbaum1960dual}, however,
can only be obtained in very special cases~\cite{klenske2016dual}.
Therefore, in most applications, \emph{exploration bonuses} are employed,
which stem from the ``optimism in the face of uncertainty'' principle~\cite{lai1985asymptotically}.
\emph{Bayesian exploration bonuses}~\cite{kolter2009near}
and other types of \emph{visitation counts}~\cite{kearns2002near} have been shown to be effective
in video games~\cite{bellemare2016unifying,ostrovski2017count}.
If the observation space is high-dimensional, exploration bonuses can be applied in the latent space.
For example, latent-space prediction error and count-based exploration
were combined in~\cite{stadie2015incentivizing}, while information gain was employed in~\cite{houthooft2016vime}.
\emph{Self-supervised prediction}~\cite{pathak2017curiosity}
and \emph{random network distillation}~\cite{burda2018exploration}
were proposed as different ways to compute the prediction error.
A comprehensive study of curiosity-driven exploration methods can be found in~\cite{burda2018large}.

Exploration bonuses are commonly added on top of a primary RL objective function to promote faster learning.
However, such approaches do not scale to the multi-task and transfer learning settings
because the knowledge gained during exploration is not reused.
In contrast, model-based approaches compress the knowledge into the model
and can later reuse it in any downstream task.

Our method can be seen as lying at the intersection
of optimal sequential experiment design and nonlinear system identification.
In the former, info-gain-maximizing strategies are well understood but for linear models;
we use these insights by treating our model as linear in the last-layer parameters.
In the latter, the focus is placed on numerical approaches and structured models
(e.g., grey-box models such as Hammerstein-Wiener model);
we use receding horizon control for numerical optimization
and basis function expansion for representing the dynamics as a black box.

We stress that trajectory optimization is essential for making the problem computationally tractable.
Approaches such as~\cite{ling2016gaussian} and~\cite{huan2016sequential}
rely on approximately solving the Bellman equation, which scales exponentially with the time horizon.
On the other hand, belief space trajectory optimization scales polynomially~\cite{patil2015scaling},
allowing for much longer horizons (e.g., we used  $150 \leq T \leq 200$,
whereas horizons of length $T \leq 4$ were considered in~\cite{ling2016gaussian}).


\section{Conclusion}
\label{sec:conclusion}
A principled algorithm for trajectory-based active exploration
in the model-based reinforcement learning setting has been proposed (Sec.~\ref{sec:rhc}).
Two acquisition functions from active learning
have been adapted to guide episodic exploration (Sec.~\ref{subsec:active-learning}):
uncertainty sampling (US, Sec.~\ref{subsec:us})
and expected variance reduction (EVR, Sec.~\ref{subsec:evr}).
Since the acquisition functions cannot be straightforwardly evaluated
due to intractability of the belief propagation over time,
an approximation has been proposed, which led to a novel algorithm,
called receding horizon curiosity (RHC, Algorithm~\ref{alg:sysid}).

The proposed RHC approach was compared to state-of-the-art
model-based and model-free exploration algorithms
on classical continuous control problems.
Empirical evaluations showed that RHC achieves higher model likelihood
and collects higher reward on downstream tasks in fewer iterations.
Although not yet real-time capable, RHC was found to be computationally faster than MAX,
thanks to being trajectory-based.
The US objective (Sec.~\ref{sec:result}) delivered a better computation/performance trade-off,
reaching the performance of EVR while being substantially easier to compute.

The experiments demonstrated that Bayesian curiosity w.r.t. last-layer parameters
interleaved with nonlinear maximum likelihood feature learning can be successfully implemented
and considerably improves exploration in low-dimensional classical control environments
even under relatively strong deterministic state propagation assumption.
Nevertheless, a number of obstacles need to be overcome to scale RHC to higher-dimensional problems.
For instance, cheaper trajectory optimization methods (e.g., first-order, Hessian-free)
could enable the use of larger number of features.
Alternatively, dimensionality reduction techniques could allow for scaling the current approach
by employing the same optimization framework but with lower-dimensional feature representations.
Finally, tractable planning methods that can utilize more expressive probabilistic models,
such as Bayesian neural networks, could allow for tackling even harder problems.



\acknowledgments{Calculations for this research were conducted
on the Lichtenberg high performance computer of the TU Darmstadt.
This project has received funding from the European Union's Horizon 2020
research and innovation programme under grant agreement  No. 640554 (SKILLS4ROBOTS).}


\small
\bibliography{lit}  

\begin{thebibliography}{42}
\providecommand{\natexlab}[1]{#1}
\providecommand{\url}[1]{\texttt{#1}}
\expandafter\ifx\csname urlstyle\endcsname\relax
  \providecommand{\doi}[1]{doi: #1}\else
  \providecommand{\doi}{doi: \begingroup \urlstyle{rm}\Url}\fi

\bibitem[Kaelbling et~al.(1998)Kaelbling, Littman, and
  Cassandra]{kaelbling1998planning}
L.~P. Kaelbling, M.~L. Littman, and A.~R. Cassandra.
\newblock Planning and acting in partially observable stochastic domains.
\newblock \emph{Artificial intelligence}, 101\penalty0 (1-2):\penalty0 99--134,
  1998.

\bibitem[Papadimitriou and Tsitsiklis(1987)]{papadimitriou1987complexity}
C.~H. Papadimitriou and J.~N. Tsitsiklis.
\newblock The complexity of markov decision processes.
\newblock \emph{Mathematics of operations research}, 12\penalty0 (3):\penalty0
  441--450, 1987.

\bibitem[Ghavamzadeh et~al.(2015)Ghavamzadeh, Mannor, Pineau, Tamar,
  et~al.]{ghavamzadeh2015bayesian}
M.~Ghavamzadeh, S.~Mannor, J.~Pineau, A.~Tamar, et~al.
\newblock Bayesian reinforcement learning: A survey.
\newblock \emph{Foundations and Trends{\textregistered} in Machine Learning},
  8\penalty0 (5-6):\penalty0 359--483, 2015.

\bibitem[Burda et~al.(2019)Burda, Edwards, Pathak, Storkey, Darrell, and
  Efros]{burda2018large}
Y.~Burda, H.~Edwards, D.~Pathak, A.~Storkey, T.~Darrell, and A.~A. Efros.
\newblock Large-scale study of curiosity-driven learning.
\newblock In \emph{ICLR}, 2019.

\bibitem[Shyam et~al.(2018)Shyam, Ja{\'s}kowski, and Gomez]{shyam2018model}
P.~Shyam, W.~Ja{\'s}kowski, and F.~Gomez.
\newblock Model-based active exploration.
\newblock \emph{arXiv preprint arXiv:1810.12162}, 2018.

\bibitem[Sun et~al.(2011)Sun, Gomez, and Schmidhuber]{sun2011planning}
Y.~Sun, F.~Gomez, and J.~Schmidhuber.
\newblock Planning to be surprised: Optimal bayesian exploration in dynamic
  environments.
\newblock In \emph{International Conference on Artificial General
  Intelligence}, pages 41--51. Springer, 2011.

\bibitem[Mehra(1974)]{mehra1974optimal}
R.~Mehra.
\newblock Optimal input signals for parameter estimation in dynamic
  systems--survey and new results.
\newblock \emph{IEEE Transactions on Automatic Control}, 19\penalty0
  (6):\penalty0 753--768, 1974.

\bibitem[Gevers and Bombois(2006)]{gevers2006input}
M.~Gevers and X.~Bombois.
\newblock Input design: From open-loop to control-oriented design.
\newblock \emph{IFAC Proceedings Volumes}, 39\penalty0 (1):\penalty0
  1329--1334, 2006.

\bibitem[Zarrop(1979)]{zarrop1979optimal}
M.~B. Zarrop.
\newblock \emph{Optimal experiment design for dynamic system identification},
  volume~21.
\newblock Springer, 1979.

\bibitem[Huan and Marzouk(2016)]{huan2016sequential}
X.~Huan and Y.~M. Marzouk.
\newblock Sequential bayesian optimal experimental design via approximate
  dynamic programming.
\newblock \emph{arXiv preprint arXiv:1604.08320}, 2016.

\bibitem[Klenske and Hennig(2016)]{klenske2016dual}
E.~D. Klenske and P.~Hennig.
\newblock Dual control for approximate bayesian reinforcement learning.
\newblock \emph{Journal of Machine Learning Research}, 17\penalty0
  (127):\penalty0 1--30, 2016.

\bibitem[Ling et~al.(2016)Ling, Low, and Jaillet]{ling2016gaussian}
C.~K. Ling, K.~H. Low, and P.~Jaillet.
\newblock Gaussian process planning with lipschitz continuous reward functions:
  Towards unifying bayesian optimization, active learning, and beyond.
\newblock In \emph{Thirtieth AAAI Conference on Artificial Intelligence}, 2016.

\bibitem[Settles(2009)]{settles2009active}
B.~Settles.
\newblock Active learning literature survey.
\newblock Technical report, University of Wisconsin-Madison Department of
  Computer Sciences, 2009.

\bibitem[Chaloner and Verdinelli(1995)]{chaloner1995bayesian}
K.~Chaloner and I.~Verdinelli.
\newblock Bayesian experimental design: A review.
\newblock \emph{Statistical Science}, pages 273--304, 1995.

\bibitem[Cohn et~al.(1996)Cohn, Ghahramani, and Jordan]{cohn1996active}
D.~A. Cohn, Z.~Ghahramani, and M.~I. Jordan.
\newblock Active learning with statistical models.
\newblock \emph{Journal of artificial intelligence research}, 4:\penalty0
  129--145, 1996.

\bibitem[Bishop(2006)]{bishop2006pattern}
C.~M. Bishop.
\newblock \emph{Pattern recognition and machine learning}.
\newblock springer, 2006.

\bibitem[Rahimi and Recht(2008)]{rahimi2008random}
A.~Rahimi and B.~Recht.
\newblock Random features for large-scale kernel machines.
\newblock In \emph{Advances in neural information processing systems}, pages
  1177--1184, 2008.

\bibitem[Rajeswaran et~al.(2017)Rajeswaran, Lowrey, Todorov, and
  Kakade]{rajeswaran2017towards}
A.~Rajeswaran, K.~Lowrey, E.~V. Todorov, and S.~M. Kakade.
\newblock Towards generalization and simplicity in continuous control.
\newblock In \emph{Advances in Neural Information Processing Systems}, pages
  6550--6561, 2017.

\bibitem[Stoer and Bulirsch(2013)]{stoer2013introduction}
J.~Stoer and R.~Bulirsch.
\newblock \emph{Introduction to numerical analysis}, volume~12.
\newblock Springer Science \& Business Media, 2013.

\bibitem[Andersson et~al.(2018)Andersson, Gillis, Horn, Rawlings, and
  Diehl]{andersson2018casadi}
J.~A. Andersson, J.~Gillis, G.~Horn, J.~B. Rawlings, and M.~Diehl.
\newblock Casadi: a software framework for nonlinear optimization and optimal
  control.
\newblock \emph{Mathematical Programming Computation}, pages 1--36, 2018.

\bibitem[Deisenroth and Rasmussen(2011)]{deisenroth2011pilco}
M.~Deisenroth and C.~E. Rasmussen.
\newblock Pilco: A model-based and data-efficient approach to policy search.
\newblock In \emph{Proceedings of the 28th International Conference on machine
  learning (ICML-11)}, pages 465--472, 2011.

\bibitem[Platt~Jr et~al.(2010)]{platt2010belief}
R.~Platt~Jr et~al.
\newblock Belief space planning assuming maximum likelihood observations.
\newblock In \emph{Proceedings of the Robotics: Science and Systems Conference,
  6th}, 2010.

\bibitem[Haarnoja et~al.(2018)Haarnoja, Zhou, Abbeel, and
  Levine]{haarnoja2018soft}
T.~Haarnoja, A.~Zhou, P.~Abbeel, and S.~Levine.
\newblock Soft actor-critic: Off-policy maximum entropy deep reinforcement
  learning with a stochastic actor.
\newblock \emph{arXiv preprint arXiv:1801.01290}, 2018.

\bibitem[Silvia(2012)]{silvia2012curiosity}
P.~J. Silvia.
\newblock Curiosity and motivation.
\newblock \emph{The Oxford handbook of human motivation}, pages 157--166, 2012.

\bibitem[Ryan and Deci(2000)]{ryan2000intrinsic}
R.~M. Ryan and E.~L. Deci.
\newblock Intrinsic and extrinsic motivations: Classic definitions and new
  directions.
\newblock \emph{Contemporary educational psychology}, 25\penalty0 (1):\penalty0
  54--67, 2000.

\bibitem[Schmidhuber(1991{\natexlab{a}})]{schmidhuber1991possibility}
J.~Schmidhuber.
\newblock A possibility for implementing curiosity and boredom in
  model-building neural controllers.
\newblock In \emph{Proc. of the international conference on simulation of
  adaptive behavior: From animals to animats}, pages 222--227,
  1991{\natexlab{a}}.

\bibitem[Schmidhuber(1991{\natexlab{b}})]{schmidhuber1991curious}
J.~Schmidhuber.
\newblock Curious model-building control systems.
\newblock In \emph{[Proceedings] 1991 IEEE International Joint Conference on
  Neural Networks}, pages 1458--1463. IEEE, 1991{\natexlab{b}}.

\bibitem[Pathak et~al.(2017)Pathak, Agrawal, Efros, and
  Darrell]{pathak2017curiosity}
D.~Pathak, P.~Agrawal, A.~A. Efros, and T.~Darrell.
\newblock Curiosity-driven exploration by self-supervised prediction.
\newblock In \emph{Proceedings of the IEEE Conference on Computer Vision and
  Pattern Recognition Workshops}, pages 16--17, 2017.

\bibitem[Lindley et~al.(1956)]{lindley1956measure}
D.~V. Lindley et~al.
\newblock On a measure of the information provided by an experiment.
\newblock \emph{The Annals of Mathematical Statistics}, 27\penalty0
  (4):\penalty0 986--1005, 1956.

\bibitem[Feldbaum(1960)]{feldbaum1960dual}
A.~Feldbaum.
\newblock Dual control theory. i.
\newblock \emph{Avtomatika i Telemekhanika}, 21\penalty0 (9):\penalty0
  1240--1249, 1960.

\bibitem[Lai and Robbins(1985)]{lai1985asymptotically}
T.~L. Lai and H.~Robbins.
\newblock Asymptotically efficient adaptive allocation rules.
\newblock \emph{Advances in applied mathematics}, 6\penalty0 (1):\penalty0
  4--22, 1985.

\bibitem[Kolter and Ng(2009)]{kolter2009near}
J.~Z. Kolter and A.~Y. Ng.
\newblock Near-bayesian exploration in polynomial time.
\newblock In \emph{Proceedings of the 26th Annual International Conference on
  Machine Learning}, pages 513--520. ACM, 2009.

\bibitem[Kearns and Singh(2002)]{kearns2002near}
M.~Kearns and S.~Singh.
\newblock Near-optimal reinforcement learning in polynomial time.
\newblock \emph{Machine learning}, 49\penalty0 (2-3):\penalty0 209--232, 2002.

\bibitem[Bellemare et~al.(2016)Bellemare, Srinivasan, Ostrovski, Schaul,
  Saxton, and Munos]{bellemare2016unifying}
M.~Bellemare, S.~Srinivasan, G.~Ostrovski, T.~Schaul, D.~Saxton, and R.~Munos.
\newblock Unifying count-based exploration and intrinsic motivation.
\newblock In \emph{Advances in Neural Information Processing Systems}, pages
  1471--1479, 2016.

\bibitem[Ostrovski et~al.(2017)Ostrovski, Bellemare, van~den Oord, and
  Munos]{ostrovski2017count}
G.~Ostrovski, M.~G. Bellemare, A.~van~den Oord, and R.~Munos.
\newblock Count-based exploration with neural density models.
\newblock In \emph{Proceedings of the 34th International Conference on Machine
  Learning-Volume 70}, pages 2721--2730. JMLR. org, 2017.

\bibitem[Stadie et~al.(2015)Stadie, Levine, and
  Abbeel]{stadie2015incentivizing}
B.~C. Stadie, S.~Levine, and P.~Abbeel.
\newblock Incentivizing exploration in reinforcement learning with deep
  predictive models.
\newblock \emph{arXiv preprint arXiv:1507.00814}, 2015.

\bibitem[Houthooft et~al.(2016)Houthooft, Chen, Duan, Schulman, De~Turck, and
  Abbeel]{houthooft2016vime}
R.~Houthooft, X.~Chen, Y.~Duan, J.~Schulman, F.~De~Turck, and P.~Abbeel.
\newblock Vime: Variational information maximizing exploration.
\newblock In \emph{Advances in Neural Information Processing Systems}, pages
  1109--1117, 2016.

\bibitem[Burda et~al.(2018)Burda, Edwards, Storkey, and
  Klimov]{burda2018exploration}
Y.~Burda, H.~Edwards, A.~Storkey, and O.~Klimov.
\newblock Exploration by random network distillation.
\newblock \emph{arXiv preprint arXiv:1810.12894}, 2018.

\bibitem[Patil et~al.(2015)Patil, Kahn, Laskey, Schulman, Goldberg, and
  Abbeel]{patil2015scaling}
S.~Patil, G.~Kahn, M.~Laskey, J.~Schulman, K.~Goldberg, and P.~Abbeel.
\newblock Scaling up gaussian belief space planning through covariance-free
  trajectory optimization and automatic differentiation.
\newblock In \emph{Algorithmic foundations of robotics XI}, pages 515--533.
  Springer, 2015.

\bibitem[Hill et~al.(2018)Hill, Raffin, Ernestus, Gleave, Traore, Dhariwal,
  Hesse, Klimov, Nichol, Plappert, Radford, Schulman, Sidor, and
  Wu]{stable-baselines}
A.~Hill, A.~Raffin, M.~Ernestus, A.~Gleave, R.~Traore, P.~Dhariwal, C.~Hesse,
  O.~Klimov, A.~Nichol, M.~Plappert, A.~Radford, J.~Schulman, S.~Sidor, and
  Y.~Wu.
\newblock Stable baselines.
\newblock \url{https://github.com/hill-a/stable-baselines}, 2018.

\bibitem[Brockman et~al.(2016)Brockman, Cheung, Pettersson, Schneider,
  Schulman, Tang, and Zaremba]{openaigym}
G.~Brockman, V.~Cheung, L.~Pettersson, J.~Schneider, J.~Schulman, J.~Tang, and
  W.~Zaremba.
\newblock Openai gym, 2016.

\bibitem[Tassa et~al.(2018)Tassa, Doron, Muldal, Erez, Li, Casas, Budden,
  Abdolmaleki, Merel, Lefrancq, et~al.]{tassa2018deepmind}
Y.~Tassa, Y.~Doron, A.~Muldal, T.~Erez, Y.~Li, D.~d.~L. Casas, D.~Budden,
  A.~Abdolmaleki, J.~Merel, A.~Lefrancq, et~al.
\newblock Deepmind control suite.
\newblock \emph{arXiv preprint arXiv:1801.00690}, 2018.

\end{thebibliography}

\clearpage
\appendix
\section{Algorithms}
Trajectory optimization for receding horizon curiosity (RHC)
is implemented using CasADi~\cite{andersson2018casadi}, a control and auto-differentiation toolbox.
The number of Fourier features in the learned dynamics model varies across environments:
MountainCar $20$, Pendulum $90$, CartPole $80$.

As a model-free baseline RL algorithm, soft actor-critic (SAC) is used~\cite{haarnoja2018soft},
as implemented in Stable Baselines~\cite{stable-baselines},
with the following parameters across all environments
\begin{align*}
\small
\gamma &= 0.99,\\
\tau &= 0.005,\\
\textrm{learning rate} &= 0.0003,\\
\textrm{buffer size} &= 50000,\\
\textrm{batch size} &= 64.
\end{align*}
The exploration bonus based on the \emph{squared prediction error} is defined as follows
\begin{align*}
r_{\mathrm{pe}}(s_t, a_t) = (s_{t+1} - \mathbb{E}_p [s_{t+1} \given s_{t}, a_{t}])^2,
\end{align*}
where $p$ denotes the model trained on the data from previous episodes.
The exploration bonus based on the \emph{information gain} is defined as the reduction of entropy
\begin{align*}
r_{\mathrm{ig}}(s_t, a_t) = \mathbb{H}(\cvec{\theta} \given \mathcal{X}_{t})
- \mathbb{H}(\cvec{\theta} \given \mathcal{X}_{t+1}),
\end{align*}
where $\mathcal{X}_t$ denotes the set of observations until time step $t$,
$\mathbb{H}$ is the entropy, and $\cvec{\theta}$ denotes the model parameters.

\section{Environments}

\paragraph{MountainCar.}
The implementation from OpenAI Gym~\cite{openaigym} is modified as follows
to accommodate the episodic exploration setting.
Car power is set to $10^{-3}$ and the speed limit is removed.
Upon reset, the car starts at the center of the valley with zero velocity.
An episode ends when the car reaches the environment bounds or after $130$ time steps.
The evaluation task is to drive the car on top of the mountain
as dictated by the stage cost $c = 10 (x - x_\textrm{goal})^2 + 0.001 a^2$,
where $x$ is the position of the car, $x_\textrm{goal}$ is the goal location, and $a$ the action.

\paragraph{Pendulum.}
The implementation from DeepMind Control Suite~\cite{tassa2018deepmind}
with observations $[\cos\theta, \sin\theta, \dot{\theta}]$
is used with the following modifications.
The pendulum is initialized handging down with zero velocity.
Each episode consists of $100$ time steps of $80\text{ms}$ duration each.
The evaluation task is to swing the pendulum up
as dictated by the stage cost $c = 100  (1 - \cos\theta) ^2 + 0.1 \sin^2\theta + 0.1 \dot{\theta}^2 + 0.001 a^2$.

\paragraph{CartPole.}
The implementation from DeepMind Control Suite~\cite{tassa2018deepmind}
with observations $[ x, \cos\theta, \sin\theta, \dot{x}, \dot{\theta} ]$ is used.
Each episode starts with the cart at the center and the pole hanging down, both having zero velocity.
The system is simulated at $50\text{Hz}$.
An episode ends when the cart reaches the state limits or after $100$ time steps.
The evaluation task is to swing the pole up,
$c = 100x^2 + 100(1 - \cos\theta)^2 + 0.1\sin^2\theta + 0.1\dot{x}^2 + 0.1\dot{\theta}^2 + 0.1a^2$.

\section{Runtimes}
Table~\ref{table:runtimes} shows how long one run of each algorithm
depicted in Fig.~\ref{fig:eval_results} on average takes.
One run consists of $20$ episodes ($x$-axis in Fig.~\ref{fig:eval_results}).
Evaluation of RHC EVR was only possible on the MountainCar environment due to its high memory demands.
Evaluations were run on a machine with an Intel Xeon E5-2670 processor.
\newcommand{\thickline}{\Xhline{2\arrayrulewidth}}
\newcolumntype{'}{!{\vrule width 1pt}}
\setlength{\tabcolsep}{10pt}
\renewcommand{\arraystretch}{1.2}
\begin{table}[b]
	\begin{center}
		\begin{tabular}{' l ' c | c | c '}
			\thickline
			& MountainCar & Pendulum & CartPole \\ \thickline
			RHC EVR & 1.51 & - & - \\ \hline
			RHC US & 0.03 & 0.80 & 0.54 \\ \hline
			SAC PE & 0.03 & 0.20 & 0.45 \\ \hline
			SAC IG & 0.03 & 0.21 & 0.48 \\ \hline
			RAND & 0.01 & 0.09 & 0.30 \\ \hline
			MAX & 9.49 & 11.17 & 5.84 \\
			\thickline
		\end{tabular}
	\end{center}
	\caption{
		Average wall-clock-time (in hours) for evaluated exploration algorithms (see Fig.~\ref{fig:eval_results}).}
	\label{table:runtimes}
\end{table}

\end{document}